\documentclass[letterpaper, 10 pt, conference]{ieeeconf}  

\usepackage[T1]{fontenc} %

\IEEEoverridecommandlockouts                              %

\usepackage{amsmath}
\usepackage{amssymb}
\usepackage{graphicx}
\graphicspath{{../figures/}}
\usepackage{booktabs}
\usepackage{makecell}
\usepackage{url}
\usepackage{cite}
\usepackage{caption}
\usepackage{placeins}
\makeatletter
\let\NAT@parse\undefined
\makeatother

\usepackage{hyperref}
\hypersetup{hidelinks}

\newcommand{\methodname}{HumanoidTTT}
\newcommand{\method}{HumanoidTTT}

\title{\LARGE \bf
\methodname: Test-Time Capability Reuse for Efficient Humanoid Control
}

\author{
    \textbf{Jingtai Yang}$^{*}$\quad
    \textbf{Yining Wu}$^{*}$\quad 
    \textbf{Yanjun Li}$^{*}$\quad 
    \textbf{Zeyu Zhang}$^{*\dag}$\quad
    \textbf{Hao Tang}$^{\ddag}$ \vspace{0.1cm}\\
    School of Computer Science, Peking University\vspace{0.05cm}\\
    \small $^*$Equal contribution. $^\dag$Project lead.
    $^\ddag$Corresponding author: bjdxtanghao@gmail.com
}

\begin{document}
\raggedbottom
\setcounter{topnumber}{1}

\makeatletter
\let\@oldmaketitle\@maketitle%
\renewcommand{\@maketitle}{\@oldmaketitle%
\vspace{0.2cm}
{\centering
\includegraphics[width=\linewidth]{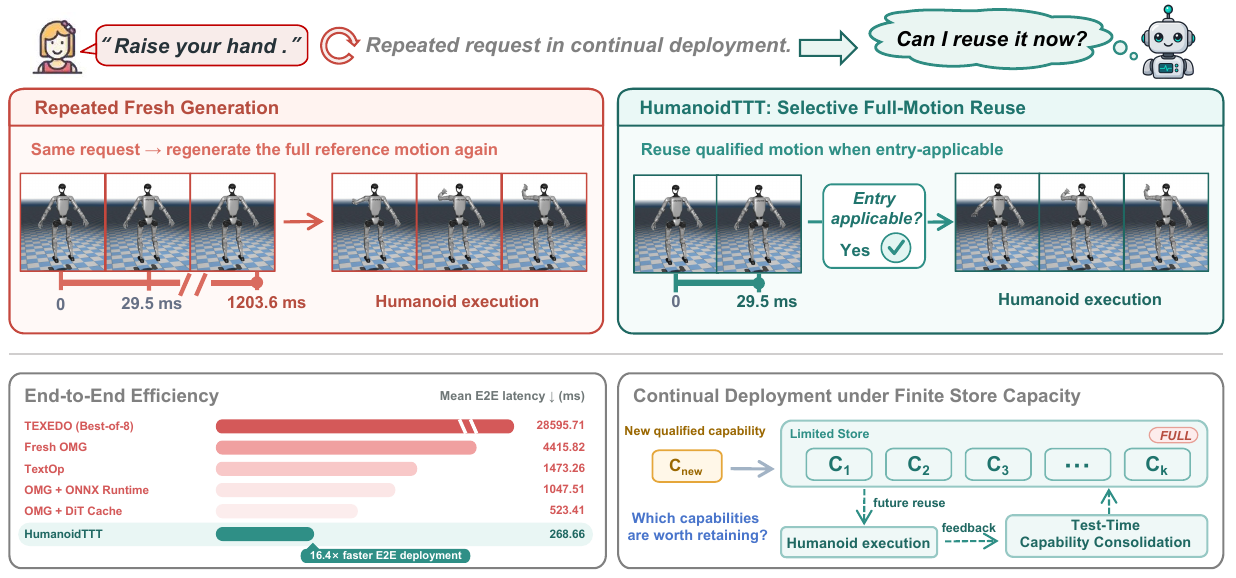}\par
\captionof{figure}{\textbf{HumanoidTTT: reusable full-motion capabilities for continual humanoid deployment.} For recurring requests, Entry Applicability enables direct reuse of qualified full motions when the current state satisfies their entry certificates, bypassing fresh generation. Test-Time Capability Consolidation uses deployment feedback to adapt the contents of a bounded Full-Motion Store. The lower-left panel compares mean end-to-end latency on the evaluated recurrent request trace.}
\label{fig:main}}
\vspace{-0.5cm}
\bigskip}%
\makeatother

\maketitle
\setcounter{figure}{1}
\thispagestyle{empty}
\pagestyle{empty}

\begin{abstract}

Recent advances in motion generation and whole-body tracking have enabled humanoid robots to execute increasingly diverse motions, yet the same motion capabilities may be requested repeatedly during continual deployment. Reliable reuse is challenging because intervening motions can change the robot's entry state, making previously successful motions unsafe to replay blindly. Meanwhile, validated capabilities accumulate during deployment, while bounded storage requires deciding which ones are worth retaining. To address these challenges, we present HumanoidTTT, a framework for test-time capability reuse in continual humanoid control. Specifically, we introduce Selective Full-Motion Reuse, which authorizes direct reuse of validated complete motions only from certified applicable entry states, allowing accepted reuse to bypass fresh generation. We further introduce Test-Time Capability Consolidation, which adapts which qualified capabilities persist in a bounded Full-Motion Store using subsequent deployment reuse as feedback. Experiments demonstrate zero unsafe accepts and a 16.4$\times$ end-to-end speedup over fresh generation, while online consolidation improves avoided generator calls by 13.2 per 200 requests over its frozen counterpart. Overall, HumanoidTTT enables reliable and efficient reuse of validated motion capabilities while adaptively retaining useful capabilities throughout continual deployment.
Code: \url{https://github.com/AIGeeksGroup/HumanoidTTT}.
Website: \url{https://aigeeksgroup.github.io/HumanoidTTT}.
\end{abstract}

\section{INTRODUCTION}
\label{sec:introduction}

Recent advances in generative motion modeling and whole-body tracking have enabled humanoid robots to synthesize and execute increasingly diverse motions from language, human-reference motion, and other conditioning signals~\cite{huang2026omg,rempe2026kimodo,chen2026holomotion}. Together, modern motion generators and tracking policies form capable generator--tracker stacks that connect motion synthesis with whole-body execution. In the continual-deployment setting considered in this work, requests for the same motion capabilities may recur over time, requiring the system to serve them multiple times over a deployment horizon.

This recurrence, however, makes reuse non-trivial. First, repeatedly invoking the full motion-generation pipeline for a motion that has already been generated and successfully executed introduces redundant computation. Directly replaying the stored motion is not a reliable alternative, because intervening motions can change the robot's entry state; matching the request identity alone therefore does not guarantee that the previously successful motion remains executable from the current state. Second, validated motions accumulate throughout deployment while the persistent store has finite capacity. Qualification establishes that a motion is eligible for reuse, but does not indicate whether it will be sufficiently useful for future requests to justify occupying a limited Store slot. Continual capability reuse therefore raises two coupled challenges: when can a stored complete motion reliably replace fresh generation, and which validated capabilities are worth retaining under finite capacity?

These two challenges suggest two corresponding design principles. To address the first, reuse should be state-dependent: a stored complete motion should replace fresh generation only when the current robot state permits its execution, rather than simply because the request identity matches. To address the second, retention should adapt to deployment demand, prioritizing qualified capabilities according to their utility for future requests rather than treating all qualified capabilities equally. Together, these principles separate continual capability reuse into a read-side applicability problem, which determines whether a stored motion can serve the current request, and a write-side retention problem, which determines which validated capabilities should persist as deployment continues.

To address the dual challenges of reliable reuse under changing entry states and capability retention under bounded storage, we develop HumanoidTTT, a framework for continual humanoid deployment that turns execution-qualified complete motions into persistent capabilities. HumanoidTTT first introduces Selective Full-Motion Reuse, where each stored capability consists of a validated complete motion and an applicability certificate defining its Entry Applicability set, i.e., the set of admissible robot entry states from which the motion can be reliably reused. For each recurring request, a matched capability is directly reused only when the current entry state falls within this set, allowing the stored complete motion to bypass the Frozen Motion Generator and proceed directly to execution; otherwise, the system falls back to fresh motion generation followed by execution-based qualification. HumanoidTTT further introduces Test-Time Capability Consolidation to manage capability persistence under a bounded Full-Motion Store. When newly qualified capabilities compete for limited storage, the consolidation policy decides whether to skip the new capability or replace an existing one, and adapts its retention decisions using subsequent reuse utility observed during deployment. This test-time adaptation changes which qualified capabilities persist in the Store, while motion generation and acceptance criteria remain fixed and each applicability certificate is frozen after acquisition. Finally, we conduct systematic experiments to evaluate HumanoidTTT from both the reuse and retention perspectives, covering full-motion reuse reliability, accepted-hit efficiency, end-to-end inference efficiency, and finite-capacity capability consolidation. HumanoidTTT reduces median preparation latency from 5057.996 ms to 29.514 ms on accepted reuse hits, corresponding to a 171.4$\times$ speedup, and achieves a 16.4$\times$ end-to-end speedup over fresh generation. Under a capacity-$10$ Full-Motion Store, online consolidation avoids 13.2 additional generator calls per 200-request stream compared with the same pretrained retention policy with online updates disabled. Together, these results demonstrate that HumanoidTTT enables reliable and efficient reuse of validated complete motions while improving capability retention under bounded deployment storage.

Our contributions to continual humanoid deployment can be summarized as follows:

\begin{itemize}
    \item we introduce Selective Full-Motion Reuse, which turns execution-qualified complete motions into persistent capabilities and enables reliable direct reuse across admissible entry states, bypassing fresh motion generation on accepted reuse hits.

   \item we introduce Test-Time Capability Consolidation, which learns which qualified capabilities should persist in a bounded Full-Motion Store based on subsequent reuse utility observed during deployment, while keeping motion generation and acceptance criteria fixed and freezing each applicability certificate after acquisition.

    \item we systematically evaluate HumanoidTTT in terms of full-motion reuse reliability, accepted-hit efficiency, end-to-end inference efficiency, and finite-capacity capability retention. HumanoidTTT achieves a 171.4$\times$ speedup in median preparation latency on accepted reuse hits and a 16.4$\times$ end-to-end speedup over fresh generation, while online consolidation avoids 13.2 additional generator calls per 200-request stream over the same pretrained retention policy without online updates.
\end{itemize}

\begin{figure*}[t]
    \centering
    \includegraphics[width=\textwidth]{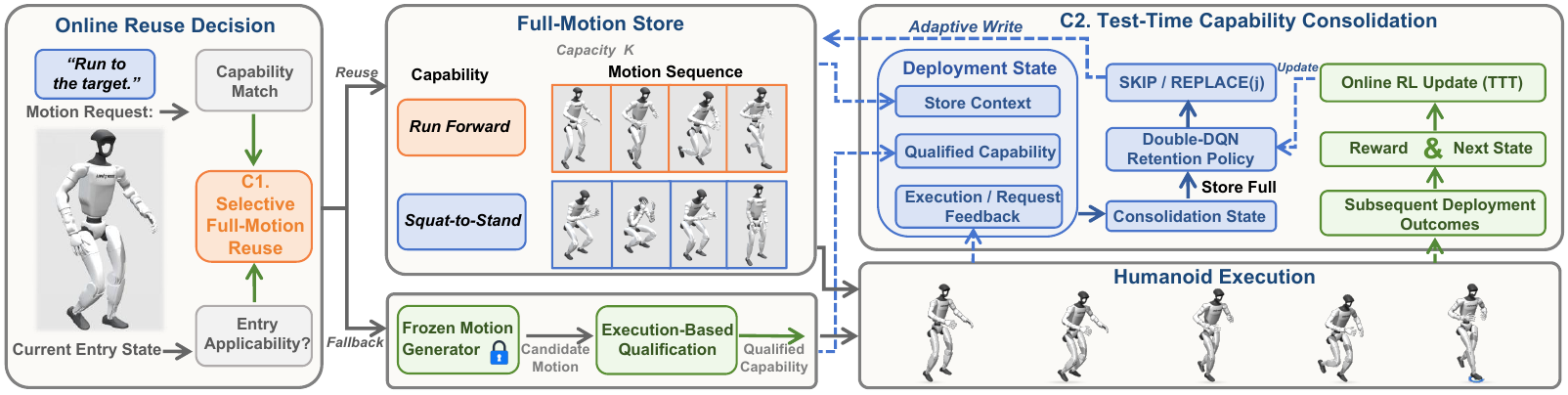}
    \caption{HumanoidTTT. Entry Applicability controls direct reuse of Validated Full-Motion Capabilities. A miss invokes the Frozen Motion Generator followed by Execution-Based Qualification. When a Qualified Capability reaches a full Store, a Store-conditioned Double-DQN chooses SKIP or replacement of an existing entry and adapts online from subsequent deployment reuse utility.}
    \label{fig:method_overview}
\end{figure*}

\section{RELATED WORK}
\label{sec:related_work}

\textbf{Humanoid Motion Generation and Execution.}
Recent advances in humanoid motion generation and execution have substantially expanded conditional motion synthesis for humans and humanoid robots. GENMO unifies human motion generation and estimation under multimodal conditions~\cite{li2025genmo}, while HY-Motion scales text-conditioned human motion generation with large-scale data and generative models~\cite{wen2025hymotion}. Kimodo further supports controllable human and humanoid motion synthesis from text and kinematic constraints~\cite{rempe2026kimodo}. Moving toward robot-native whole-body control, OMG generates humanoid motions from language, audio, and human-reference conditions~\cite{huang2026omg}, while HoloMotion provides a general whole-body tracking policy for executing diverse reference motions~\cite{chen2026holomotion}. Together, these works provide increasingly capable motion-generation and tracking stacks, but primarily focus on producing or executing new reference motions. HumanoidTTT instead builds on an existing generator--tracker stack and studies how previously qualified complete motions can persist and be reused during continual deployment.

\textbf{Memory, Caching, and Executable Behavior Reuse in Robotics.}
Prior work on memory, caching, and executable behavior reuse has explored reuse at different levels of robotic inference and behavior. MemER retrieves task-relevant keyframes from long experience histories to condition long-horizon control~\cite{sridhar2026memer}, while VLA-Cache reuses cached representations for minimally changed visual tokens across consecutive observations~\cite{xu2025vlacache}. ActionCache caches intermediate actions to warm-start iterative action generation, reducing generation cost while retaining a refinement stage~\cite{oi2026actioncache}. RT-Cache moves closer to direct behavior reuse by retrieving and replaying short multi-step motion snippets from previously successful trajectories~\cite{kwon2025rtcache}. CacheMPC further separates retrieval from acceptance, reusing finite-horizon control solutions only when a per-query certificate verifies their admissibility~\cite{khandelwal2026cachempc}. These works demonstrate reuse from intermediate computation to executable control, while also highlighting that retrieval alone does not guarantee executability. HumanoidTTT instead targets execution-qualified complete humanoid motions: a matched capability is directly reused only when the current robot state belongs to its certified Entry Applicability set, allowing an accepted reuse hit to bypass the Frozen Motion Generator entirely.

\textbf{Test-Time Adaptation and Capability Retention.}
Robot test-time adaptation uses deployment experience to modify how pretrained policies produce behavior through lightweight adaptation interfaces. TTT-VLA optimizes a latent prompt while leaving the underlying policy unchanged~\cite{zhang2026tttvla}, RoboTTT encodes long-horizon deployment context into fast weights updated during inference~\cite{jiang2026robottt}, and ZPRL performs online reinforcement learning through latent perturbations while freezing the base policy~\cite{yu2026zprl}. In parallel, cache-management research has studied retention under bounded capacity, ranging from fixed eviction heuristics such as LRU and LFU to learned replacement policies such as LeCaR and Learning Relaxed Belady~\cite{vietri2018lecar,song2020lrb}. HumanoidTTT differs in its adaptation target: motion generation and acceptance criteria remain fixed, each applicability certificate is frozen after acquisition, and online learning adapts which qualified full-motion capabilities persist in a bounded Full-Motion Store. By updating retention decisions from subsequently observed Future Reuse Utility, test-time adaptation changes what the deployed humanoid retains for future generator-free service rather than how new motions are produced.

\section{METHOD: HUMANOIDTTT}
\label{sec:method} 

\subsection{Overview}

HumanoidTTT instantiates Test-Time Capability Learning by maintaining a persistent Full-Motion Store during continual deployment. Fig.~\ref{fig:method_overview} summarizes its read, acquisition, and consolidation paths. Entry Applicability selects validated complete motions for direct execution from changed robot states. Reuse misses invoke a Frozen Motion Generator followed by fixed execution-based qualification. Qualified arrivals enter available capacity directly, whereas competition for a full Store invokes a compact Store-conditioned Double-DQN that adapts writeback decisions from causally observed subsequent reuse.

\suppressfloats[t]
\begin{figure}[t]
\captionsetup{skip=5pt}

\vspace{0pt}
\centering
\parbox[t]{\linewidth}{\vspace{0pt}\includegraphics[width=\linewidth]{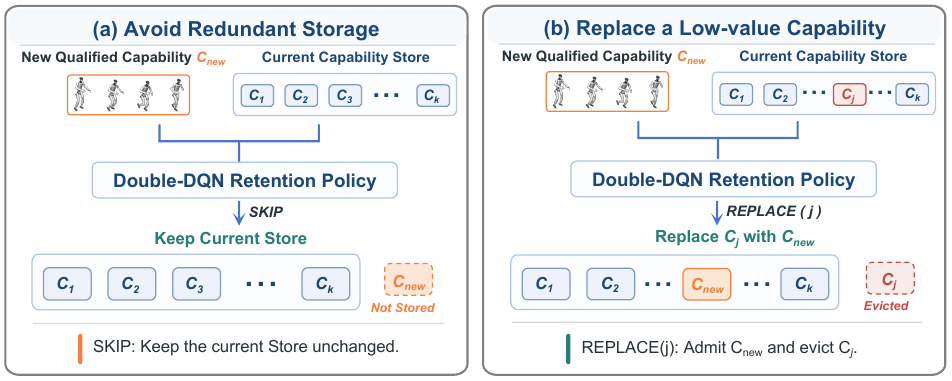}}
\caption{Consolidation at a qualified full-Store miss. (a) SKIP retains the current Store without admitting the arrival. (b) REPLACE($j$) admits the new capability and evicts entry $j$. The Double-DQN scorer selects an action using Store-conditioned features.}
\label{fig:consolidation_decisions}
\end{figure}
\begin{figure}[t]
\captionsetup{skip=5pt}

\vspace{0pt}
\centering
\parbox[t]{\linewidth}{\vspace{0pt}%
{\fontsize{7.5}{8}\selectfont\bfseries\noindent\hspace*{0.061\linewidth}(a)\hspace{0.46\linewidth}(b)\par}\vspace{2pt}
\includegraphics[width=\linewidth,trim=132bp 12bp 0 12bp,clip]{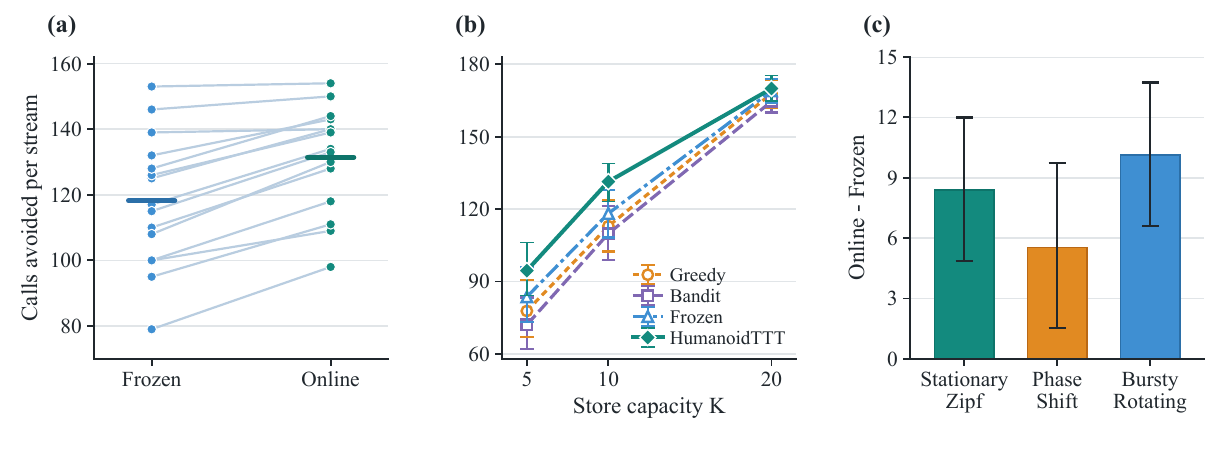}}
\caption{Capacity and request dynamics. (a) Calls avoided versus Store capacity. (b) Online-minus-frozen gains across three request families. Error bars show stream-level 95\% bootstrap confidence intervals.}
\label{fig:online_capacity_workload}

\end{figure}

\subsection{Problem Formulation}
\label{sec:problem_formulation}

At deployment step $t$, the system receives a motion request $q_t$ and observes the robot state $s_t$ with the required execution history $H_t$. For a candidate or stored complete motion $\tau_i$, a fixed entry representation produces
\begin{equation}
    x_t^{(i)}=f_{\mathrm{ent}}(s_t,H_t,\tau_i)\in\mathbb{R}^{45}.
    \label{eq:entry_descriptor}
\end{equation}
The descriptor combines candidate-relative joint and root offsets, joint and root velocity summaries, foot-contact states, physical margins, and short execution-history statistics. The capacity-$K$ Full-Motion Store is
\begin{equation}
    c_i=(d_i,\tau_i,C_i),\quad
    \mathcal{M}_t=\{c_i\}_{i=1}^{n_t},\quad n_t\leq K,
    \label{eq:full_motion_store}
\end{equation}
where $d_i$ is the capability identity and $C_i$ is a fixed candidate-specific certificate recording the applicability set $A_i$. Deployment-time learning adapts only capability consolidation; entry representation, acceptance criteria, and motion generation remain fixed.

\subsection{Selective Full-Motion Reuse}
\label{sec:selective_full_motion_reuse}

Capability Match queries $\mathcal{M}_t$ for an exact capability/signature match to $q_t$. Reuse requires fixed physical checks and candidate-relative certificate membership:
\begingroup
\thinmuskip=1mu\medmuskip=1mu\thickmuskip=1mu
\begin{equation}
\operatorname{Accept}_t(i)=\operatorname{Match}(q_t,d_i)\land\operatorname{Phys}(s_t,H_t)\land\bigl(x_t^{(i)}\in A_i\bigr).
\label{eq:entry_applicability}
\end{equation}
\endgroup
$\operatorname{Phys}$ denotes the fixed finite-state, stability, contact, penetration, joint-limit, and history checks. An accepted $\tau_i$ is retrieved for direct execution without motion generation.

Each local cell is centered at $\mu_i=x_{t_i}^{(i)}$, recorded at the start of the capability's first successful deployment execution. The shared scale vector $s$ is frozen from an independent rule-development split using robust per-dimension statistics, and the radius is fixed at $r=2.35$. For the 42 non-exact dimensions $\mathcal{C}$, the candidate-specific applicability set is
\begin{equation}
A_i=\left\{x\in\mathbb{R}^{45}:|x_k-\mu_{i,k}|\leq r s_k,\ \forall k\in\mathcal{C}\right\}.
\label{eq:local_applicability}
\end{equation}
Membership additionally requires the exact contact/history conditions $c_L=c_R=1$ and $n_{\mathrm{history}}=32$.

On a Store miss, the Frozen Motion Generator produces a Candidate Motion. Its 45-D candidate-relative entry descriptor is recorded before the fixed execution stack executes it. If this deployment execution satisfies the fixed success and physical criteria, the recorded descriptor becomes the center of $A_i$, and the motion becomes a Qualified Capability eligible for writeback. Failed candidates are not stored. The resulting applicability certificate remains fixed after acquisition.

\subsection{Test-Time Capability Consolidation}
\label{sec:test_time_capability_consolidation}

Test-Time Capability Consolidation is formulated as an online reinforcement learning problem over capacity-constrained Store writeback. As Qualified Capabilities accumulate, the consolidation policy learns which capabilities should persist in the bounded Full-Motion Store from reuse utility observed over subsequent deployment requests. The consolidation policy is invoked exactly when a qualified miss creates competition for a full Store, expressed by $\chi_t=\mathbf{1}[\mathrm{miss}_t\land\mathrm{qualified}_t\land|\mathcal{M}_t|=K]$. Qualification failure ends acquisition without writeback; available capacity receives a Qualified Capability through direct INSERT. At a trigger, the policy constructs one Store-conditioned feature vector $\phi_t^a\in\mathbb{R}^{71}$ for each candidate action $a\in\mathcal{A}_t$. A shared action scorer evaluates these action-specific representations. Its actions and Store transition are
\begin{equation}
\begin{aligned}
    \mathcal{A}_t
        &=\{\mathrm{SKIP},\mathrm{REPLACE}(1),\ldots,\mathrm{REPLACE}(K)\},\\
    \mathcal{M}_{t+1}
        &=\begin{cases}
            \mathcal{M}_t, & a_t=\mathrm{SKIP},\\
            (\mathcal{M}_t\setminus\{c_j\})\cup\{c_{\mathrm{new}}\},
                & a_t=\mathrm{REPLACE}(j).
        \end{cases}
\end{aligned}
\label{eq:store_transition}
\end{equation}
SKIP preserves the current Store. $\mathrm{REPLACE}(j)$ admits the incoming capability and removes entry $j$. The reward is $r_t=\text{execution-successful Store reuses}/\text{observed requests}$.

The consolidation policy is initialized from a fixed pretrained checkpoint and continues to adapt through online reinforcement learning during deployment. Whenever a qualified miss arrives at a full Store, the online network selects either SKIP or $\mathrm{REPLACE}(j)$ from the current Store-conditioned state. After the writeback decision, the system accumulates execution-successful Store reuse over subsequent requests. When the next qualified full-Store miss occurs, the previous transition is closed, and the reuse fraction observed over the intervening request interval is assigned as its reward. This transition is then used to perform a Double-DQN gradient update on the online network, while the target network is synchronized periodically. During this online reinforcement learning process, motion generation, entry acceptance, and qualification remain fixed, while only the Store consolidation policy is updated from causally observed reuse utility. Double-DQN uses the online network to select the next action and the target network to evaluate it:
\begin{equation}
\begin{aligned}
    a^\star
        &=\arg\max_{a\in\mathcal{A}_{t+1}}Q_\theta(\phi_{t+1}^{a}),\qquad
    y_t=r_t+\gamma Q_{\bar{\theta}}(\phi_{t+1}^{a^\star}),\\
    \mathcal{L}(\theta)
        &=\operatorname{Huber}\!\left(Q_\theta(\phi_t^{a_t})-y_t\right).
\end{aligned}
\label{eq:double_dqn}
\end{equation}
The online network minimizes Huber loss with Adam. A compact shared action scorer implements each action value with layers $71\!\rightarrow\!128\!\rightarrow\!64\!\rightarrow\!1$.

Fig.~\ref{fig:consolidation_decisions} illustrates SKIP, which preserves the current Store, and REPLACE, which admits a qualified arrival by evicting one stored capability. Both operate within the fixed capacity.

\subsection{Kernel-Aware Real-Time Deployment}
\label{sec:efficient_deployment}

HumanoidTTT uses a heterogeneous CPU--GPU execution path that separates frequent lightweight capability management from compute-intensive motion generation. For accepted reuse requests, Full-Motion Store lookup, Entry Applicability evaluation, complete-motion retrieval, and Store writeback remain on the CPU without activating the motion generator. Capacity-constrained test-time consolidation also runs in CPU FP32, where the compact Double-DQN is invoked only when a qualified capability competes for a full Store. This event-driven execution keeps online adaptation off the critical GPU generation path and avoids making test-time learning a fixed cost for every request.

Reuse misses activate the GPU generation path. Text encoding, history and motion-representation processing, and decoding use PyTorch CUDA, while diffusion denoising is dispatched through the ONNX Runtime CUDA Execution Provider~\cite{onnxruntime} using optimized CUDA/cuBLAS-backed runtime kernels with the standard CPU fallback provider registered. Keeping the small and sparsely invoked consolidation network on the CPU avoids unnecessary GPU kernel launches, device synchronization, and cross-device transfer for lightweight updates. Conditional generator invocation and this kernel-aware CPU--GPU scheduling jointly reduce critical-path computation for recurrent requests and support low end-to-end deployment latency.

\begin{figure*}[!t]
\centering
\includegraphics[width=\textwidth]{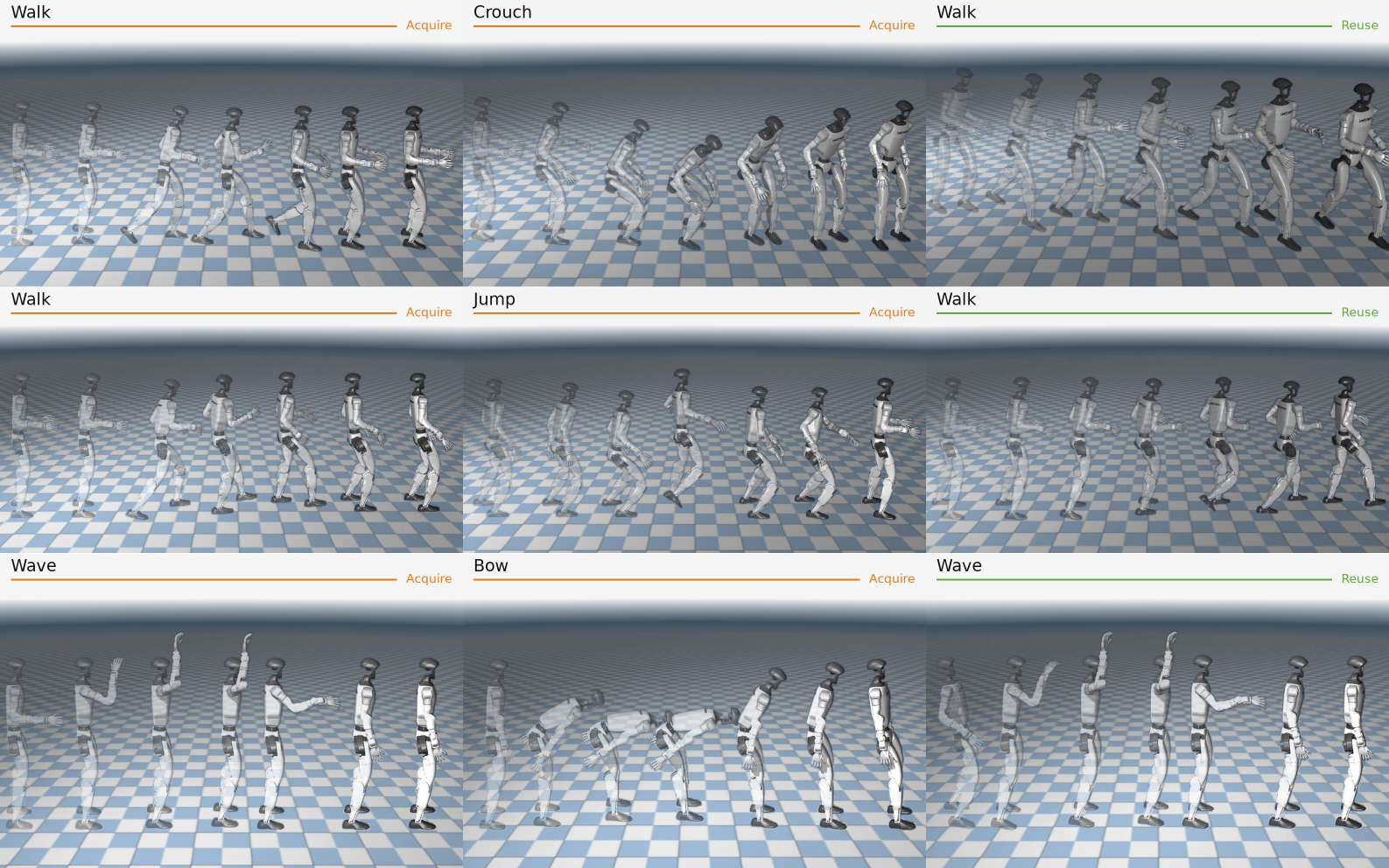}
\caption{Full-Motion Capability Reuse across Different Entry States in Continual Deployment. Each row follows a continuous acquire A, acquire B, reuse A sequence. The intervening motion changes the robot state before A is requested again, so reuse begins from a changed valid entry state. When Entry Applicability is satisfied, \method{} reuses the previously acquired complete motion across these changed entry states.}
\label{fig:persistent_capabilities}
\end{figure*}

\begin{figure*}[!t]
\centering
\includegraphics[width=\textwidth]{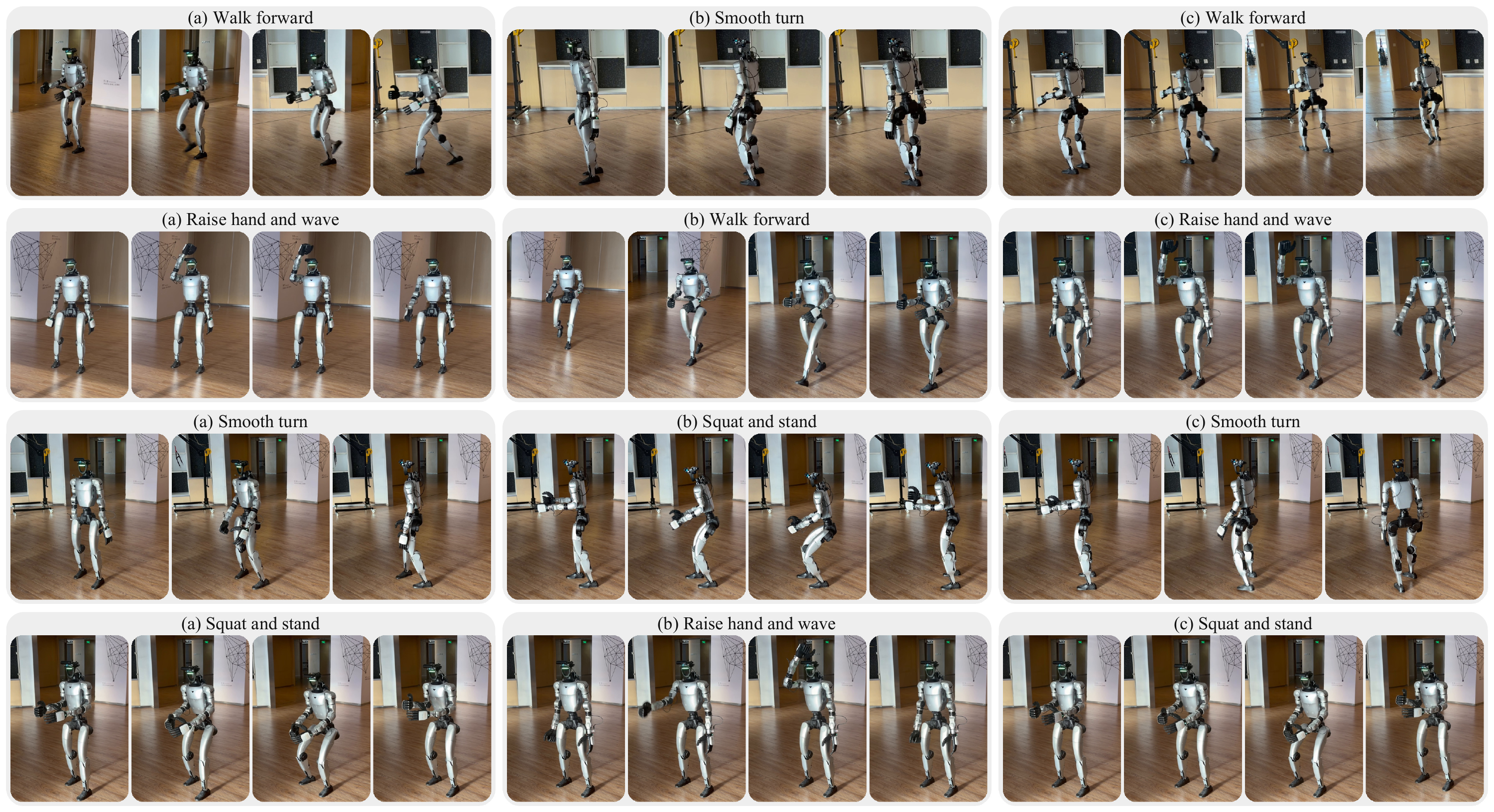}
\caption{Full-motion reuse during real-world continual deployment. Each row shows an A--B--A execution sequence: (a) initial execution of motion A; (b) execution of motion B; and (c) reuse of motion A when requested again. The four sequences cover walking forward, turning, waving, and squatting and standing, illustrating reuse after different intervening motions.}
\label{fig:real_aba}
\end{figure*}

\setcounter{topnumber}{2}
\section{EXPERIMENTS}
\label{sec:experiments}

We evaluate \method{} along four dimensions: selective full-motion reuse reliability, accepted-reuse preparation efficiency, end-to-end inference efficiency, and finite-capacity capability consolidation.

\subsection{Setup}
\label{sec:exp_setup}

We use a frozen motion generator and a fixed execution stack. A stored capability consists of a validated complete G1 motion together with the entry information used to determine whether that motion remains executable from the current state. The generator stays frozen throughout all experiments in this section; deployment-time learning acts only on capability consolidation.

Real-world experiments are conducted on a Unitree G1 humanoid using the HoloMotion whole-body tracker~\cite{chen2026holomotion} to execute motion references. We use A--B--A request sequences to demonstrate full-motion reuse during continuous execution.

Motivated by prior evaluations of humanoid tracking and execution quality~\cite{huang2026omg,chen2026holomotion}, we compare the generator baselines, TextOp~\cite{xie2026textop}, and TEXEDO~\cite{cao2026texedo} on the same fixed 200-request stream under a common G1 deployment protocol. Physical success is computed over all requests, whereas fall and joint-limit violation rates use executed requests. Tracking MPJPE includes observed prefixes of failed executions and is averaged across executed requests.

For selective reuse, an independent rule-development split fixes the entry-feature scaling, local-cell radius, and acceptance criteria. Successful deployment executions instantiate candidate-specific applicability sets. Evaluation uses 200 held-out formal motion endpoints and 887 OOD entry conditions. Safe and unsafe outcomes are derived from the frozen MuJoCo replay protocol. We report accepted coverage, unsafe accepts, OOD false accepts, and accepted-path replay success.

For finite-capacity consolidation, we use controlled identity-level replay over 15 matched 200-request streams. All methods share the request streams, initial Store, qualified candidate ledger, and frozen execution-derived qualification, applicability, and fallback outcomes; only the consolidation policy changes. Unless stated otherwise, the Store capacity is $K=10$, and each complete 200-request stream is one statistical unit.

\begin{table}[t]
\caption{Motion execution quality under a common deployment protocol. All methods use the same 200-request stream. Success and joint-limit violation rates are reported as percentages; MPJPE is in millimeters. Success is measured over all requests; joint-limit violation rates use executed requests. Tracking MPJPE includes observed prefixes of failed executions. Safety termination may precede a fall. Bold and underlining denote the best and second-best results, respectively.}
\label{tab:execution_quality}
\centering
\setlength{\tabcolsep}{6pt}
\renewcommand{\arraystretch}{1.15}
\small
\resizebox{\linewidth}{!}{%
\begin{tabular}{@{}lrrr@{}}
\toprule
Method & \begin{tabular}[b]{@{}r@{}}Physical \\ success $\uparrow$\end{tabular} & \begin{tabular}[b]{@{}r@{}}Tracking \\ MPJPE $\downarrow$\end{tabular} & \begin{tabular}[b]{@{}r@{}}Joint-limit \\ violation $\downarrow$\end{tabular} \\
\midrule
Fresh OMG~\cite{huang2026omg} & 80.50 & 21.24 & 5.15 \\
OMG + DiT Cache~\cite{liu2025teacache} & \underline{82.00} & \textbf{20.62} & 5.15 \\
TextOp~\cite{xie2026textop} (G1 adaptation) & 62.00 & 23.07 & 26.94 \\
TEXEDO~\cite{cao2026texedo} + HoloMotion~\cite{chen2026holomotion} ($N=32$) & 53.00 & 43.31 & 15.96 \\
TEXEDO~\cite{cao2026texedo} + SONIC~\cite{luo2026sonic} ($N=32$) & 58.50 & 32.46 & \textbf{1.59} \\
\midrule
\textbf{\method{} (Ours)} & \textbf{83.00} & \underline{20.94} & \underline{3.08} \\
\bottomrule
\end{tabular}%
}
\end{table}

\subsection{Execution Quality and Reuse Reliability}
\label{sec:exp_reuse}

Table~\ref{tab:execution_quality} compares motion execution quality. \method{} achieves 83.0\% physical success, compared with 80.5\% for Fresh OMG~\cite{huang2026omg} and 82.0\% for OMG + DiT Cache~\cite{liu2025teacache}. Its tracking MPJPE is 20.94\,mm, close to both baselines, while its joint-limit violation rate is 3.08\%, below their 5.15\%. These results indicate that full-motion reuse maintains competitive execution quality on this request stream. We then use independent entry-state tests to further examine the reliability of reuse decisions.

We evaluate the executability of previously validated complete motions from the robot's current state. Table~\ref{tab:reuse_reliability} shows that acceptance alone is not enough. Task-ID replay accepts all 200 endpoints, but only 249/600 accepted replays succeed and all 887 OOD conditions are falsely accepted. Nearest-entry retrieval preserves substantial acceptance, yet about 70\% of its accepted endpoints are unsafe under the recorded applicability labels.

\method{} accepts 180/200 endpoints, and every accepted endpoint is labeled safe. No unsafe accept and no OOD false accept is observed, and all 540 accepted replays succeed. These results show that Entry Applicability preserves broad reuse coverage while maintaining a reliable accepted path.

\begin{table}[!t]
\caption{Selective full-motion reuse reliability. Safe acceptance is evaluated per endpoint: an endpoint is safe for a method only when all three replays of that method's accepted path satisfy the fixed criterion. Unsafe accepts are accepted endpoints that fail this criterion. Replay success and catastrophic failures are counted per rollout; Cat.\ denotes catastrophic rollouts. OOD false accepts are evaluated on 887 entry conditions.}
\label{tab:reuse_reliability}
\centering
\setlength{\tabcolsep}{2.7pt}
\renewcommand{\arraystretch}{1.35}
\scriptsize
\resizebox{\linewidth}{!}{
\begin{tabular}{lrrrrrr}
\toprule
Method & \makecell{Accepted\\coverage $\uparrow$} & \makecell{Safe\\accepted $\uparrow$} & \makecell{Unsafe\\accepts $\downarrow$} & \makecell{OOD false\\accepts $\downarrow$} & \makecell{Replay\\success $\uparrow$} & \makecell{Cat.\\$\downarrow$} \\
\midrule
Task-ID Replay & 200/200 & 58/200 & 142 & 887/887 & 249/600 & 1 \\
Global Nearest-Entry & 178/200 & 54/200 & 124 & 274/887 & 232/534 & 1 \\
Per-Motion Nearest-Entry & 172/200 & 52/200 & 120 & 277/887 & 222/516 & 1 \\
\method{} (Ours) & 180/200 & 180/200 & 0 & 0/887 & 540/540 & 0 \\
\bottomrule
\end{tabular}
}
\end{table}

Fig.~\ref{fig:persistent_capabilities} visualizes full-motion capability acquisition and reuse during continual deployment. Each row shows a continuous execution sequence in which the robot first acquires one complete motion, then performs a different intervening motion that changes its current state, and subsequently reuses the previously acquired motion when the request recurs. Reuse can start from an entry state different from that observed during the first execution, provided that the resulting state remains entry-applicable. The stored complete motion can then be reused reliably from this different valid entry state. This qualitative behavior is consistent with the accepted-path reliability results in Table~\ref{tab:reuse_reliability}.

We further demonstrate A--B--A execution sequences on the physical robot in Fig.~\ref{fig:real_aba}. Four sequences combine walking forward, turning, waving, and squatting and standing. After executing motion A and an intervening motion B, the robot reuses the stored complete motion when A is requested again and Entry Applicability is satisfied. These real-world examples complement the simulation sequences in Fig.~\ref{fig:persistent_capabilities} and illustrate motion reuse on physical hardware.

\subsection{Accepted-Reuse Preparation Efficiency}
\label{sec:exp_hit_efficiency}

Once Entry Applicability authorizes reuse, \method{} directly reads the validated complete motion and bypasses generation. Table~\ref{tab:hit_latency} isolates this local mechanism on confirmed cache-hit contexts. The measured path contains the entry gate and stored-motion read; first acquisition, misses, fallback, and the broader consolidation lifecycle are outside this timing boundary.

Median preparation latency drops from 5057.996\,ms for a fresh generator call to 29.514\,ms for accepted full-motion reuse, a $171.4\times$ speedup. The accepted path is also $18.4\times$ faster at the median than the generator-side DiT-cache variant, while maintaining a 63.115\,ms P95.

\subsection{End-to-End Inference Efficiency}
\label{sec:exp_e2e}

We next evaluate the full path from request arrival to an executable G1 reference in FP32 on a 200-request high-recurrence trace fixed before timing and shared by all methods. This E2E trace is separate from the matched finite-capacity streams used in Table~\ref{tab:consolidation_main}. Table~\ref{tab:e2e_efficiency} shows that \method{} achieves the lowest mean end-to-end latency in this comparison, 268.66\,ms: $16.4\times$ faster than the fresh generator, $3.90\times$ faster than the ONNX-optimized generator path, and $1.95\times$ faster than the DiT-cache pipeline.

\method{} averages 75.46 GFLOPs per request, 89.0\% below the fresh generator and 72.8\% below the DiT-cache pipeline, while peak VRAM remains close to the optimized generator variants. Tables~\ref{tab:hit_latency} and~\ref{tab:e2e_efficiency} together show that direct reuse is inexpensive on a confirmed hit and that the complete inference system retains this advantage on the evaluated recurrent trace.

\begin{table}[!t]
\centering
\caption{Accepted-reuse preparation latency on confirmed hit contexts. Latency includes Entry Applicability and stored-motion retrieval. Lower is better.}
\label{tab:hit_latency}
\setlength{\tabcolsep}{3pt}
\renewcommand{\arraystretch}{1.35}
\scriptsize
\resizebox{\linewidth}{!}{%
\begin{tabular}{@{}lrrr@{}}
\toprule
Method & Mean $\downarrow$ & P50 $\downarrow$ & P95 $\downarrow$ \\
\midrule
Fresh generator~\cite{huang2026omg}
& 5092.975 & 5057.996 & 5491.174 \\
Generator + ONNX Runtime~\cite{onnxruntime}
& 1148.270 & 1123.820 & 1235.144 \\
Generator + DiT Cache~\cite{liu2025teacache}
& 558.916 & 543.084 & 629.073 \\
\method{} (Ours)
& 38.409 & 29.514 & 63.115 \\
\bottomrule
\end{tabular}%
}
\end{table}

\begin{table}[!t]
\centering
\caption{End-to-end inference efficiency in FP32 on the fixed 200-request high-recurrence trace. Latency spans request arrival to an executable G1 reference. Lower is better.}
\label{tab:e2e_efficiency}
\setlength{\tabcolsep}{3pt}
\renewcommand{\arraystretch}{1.25}
\scriptsize
\resizebox{\linewidth}{!}{%
\begin{tabular}{@{}lrrr@{}}
\toprule
Method
& \makecell{Mean E2E\\latency $\downarrow$}
& \makecell{Average\\compute $\downarrow$}
& \makecell{Peak\\memory $\downarrow$} \\
\midrule
Fresh generator~\cite{huang2026omg}
& 4415.82 & 685.97 & 1.529 \\
Generator + ONNX Runtime~\cite{onnxruntime}
& 1047.51 & 685.97 & 1.593 \\
Generator + DiT Cache~\cite{liu2025teacache}
& 523.41 & 277.44 & 1.593 \\
TextOp~\cite{xie2026textop}
& 1473.26 & 34.11 & 1.934 \\
TEXEDO~\cite{cao2026texedo} (Best-of-8)
& 28595.71 & 243.74 & 2.617 \\
\method{} (Ours)
& 268.66 & 75.46 & 1.595 \\
\bottomrule
\end{tabular}%
}
\end{table}

\subsection{Finite-Capacity Capability Consolidation}
\label{sec:exp_consolidation}

The read path becomes more valuable when the Store keeps the capabilities that are likely to be requested again. We therefore compare six consolidation policies under a controlled $K=10$ Store. Every method sees the same request streams and qualified candidates and uses the same applicability gate and frozen-generator fallback; the only difference is the retention policy.

Table~\ref{tab:consolidation_main} shows that \method{} avoids 131.4 generator calls per 200-request stream and reduces the mean generator-call count to 68.6. It improves over LRU by 33.6 calls avoided, over one-step greedy replacement by 18.467, and over the identical pretrained RL policy with updates disabled by 13.2. These gains show that write-side decisions strongly influence how much future generator-free service a finite Store can provide.

\begin{table}[!t]
\caption{Finite-capacity capability consolidation at $K=10$. Results are means over 15 matched 200-request streams; all components except the consolidation policy are held fixed.}
\label{tab:consolidation_main}
\centering
\setlength{\tabcolsep}{4.2pt}
\renewcommand{\arraystretch}{1.1}
\small
\begin{tabular*}{\linewidth}{@{\extracolsep{\fill}}lrrrr@{}}
\toprule
Method & \makecell{Calls\\avoided $\uparrow$} & \makecell{SD} & \makecell{Avoided/\\slot $\uparrow$} & \makecell{Generator\\calls $\downarrow$} \\
\midrule
LRU & 97.800 & 22.904 & 9.780 & 102.200 \\
LFU & 110.467 & 21.709 & 11.047 & 89.533 \\
One-Step Greedy & 112.933 & 21.697 & 11.293 & 87.067 \\
Contextual Bandit & 109.867 & 23.130 & 10.987 & 90.133 \\
Pretrained RL, Frozen & 118.200 & 20.232 & 11.820 & 81.800 \\
\method{} (Ours) & 131.400 & 16.017 & 13.140 & 68.600 \\
\bottomrule
\end{tabular*}
\end{table}

\subsection{Effects of Capacity and Request Dynamics}
\label{sec:exp_when}

Fig.~\ref{fig:online_capacity_workload}(a) shows that the online advantage is largest when capacity is tight: relative to the same frozen checkpoint, online consolidation adds 10.867 calls avoided at $K=5$ and 13.2 at $K=10$, while the gap contracts to 1.0 call at $K=20$. Fig.~\ref{fig:online_capacity_workload}(b) shows positive mean online-minus-frozen gains under Stationary Zipf, Phase Shift, and Bursty Rotating request streams. The aggregate $K=10$ comparison is reported in Table~\ref{tab:consolidation_main}.

\section{LIMITATIONS AND FUTURE WORK}

HumanoidTTT currently freezes each Entry Applicability certificate after capability acquisition, leaving persistent changes in robot dynamics, environmental conditions, or admissible entry regions outside the current adaptation scope. Future work will extend Entry Applicability to update from deployment experience and evaluate the resulting capability reuse and finite-capacity consolidation over longer and more diverse real-world deployments.

\section{CONCLUSION}
\label{sec:conclusion}

In this work, we presented HumanoidTTT, a framework for test-time capability reuse toward efficient humanoid control in continual deployment. HumanoidTTT couples two complementary mechanisms: Selective Full-Motion Reuse enables reliable direct reuse of validated complete motions across applicable entry states, while Test-Time Capability Consolidation adaptively retains useful capabilities under bounded storage. Experiments demonstrate zero unsafe accepts, a 16.4$\times$ end-to-end speedup over fresh generation, and improved generator-call avoidance through online consolidation. Overall, HumanoidTTT shows that validated complete motions can be reliably reused and adaptively retained as persistent capabilities for efficient continual humanoid control.

\section*{ACKNOWLEDGMENT}

Generative artificial intelligence tools, including OpenAI ChatGPT and Codex, were used to assist with manuscript organization, drafting and language editing, as well as the design and refinement of scientific figures. All technical claims, equations, experimental protocols, numerical results, citations, and final manuscript content were reviewed and verified by the authors.

\let\originalthebibliography\thebibliography
\renewcommand{\thebibliography}[1]{\originalthebibliography{#1}\interlinepenalty=0\clubpenalty=10000\widowpenalty=10000}
\bibliographystyle{IEEEtran}
\bibliography{ICRA2027_conference}

\end{document}